\documentclass{article}

\PassOptionsToPackage{numbers, compress}{natbib}
 \usepackage[preprint]{neurips_2026}

\usepackage[utf8]{inputenc} 
\usepackage[T1]{fontenc}    
\usepackage{hyperref}       
\usepackage{url}            
\usepackage{booktabs}       
\usepackage{amsfonts}       
\usepackage{nicefrac}       
\usepackage{microtype}      

\usepackage{url}
\usepackage{amsmath}
\usepackage{amsfonts}
\usepackage{booktabs} 
\usepackage{multicol}
\usepackage{multirow}
\usepackage{placeins}
\usepackage{marvosym}
\usepackage{wrapfig}
\usepackage{algorithm}
\usepackage{courier}
\usepackage{graphicx}
\usepackage[table]{xcolor}
\usepackage{color,colortbl}
\usepackage{amssymb}

\title{One Adapter, Many Tasks: Task-Conditioned Feature Transformations for Continual Learning}

\author{%
  Yunxiang Fu \quad \quad
  Meng Lou \quad \quad
  Yizhou Yu \\
  School of Computing and Data Science \\
  The University of Hong Kong 
}

\begin{document}

\maketitle

\begin{abstract}

Class-incremental learning (CIL) requires a model to incrementally learn tasks that contain new classes without accessing earlier training data while preserving the ability to recognize all seen classes. 
Recently, pretrained-model-based approaches have become prevalent by adapting a frozen backbone with additional lightweight trainable modules.
Existing methods, however, exhibit limitations: task-specific adapters learn explicit per-task representations but are parameter- and computation-inefficient, while LoRA-based merging methods combine per-task LoRA parameters into a single model whose static aggregated weights cause representation interference during inference.
To address these problems, we present \textbf{FACET}: task-conditioned \textbf{F}e\textbf{A}ture transformation with \textbf{C}ondition\textbf{E}d feature consis\textbf{T}ency, achieving excellent parameter efficiency while producing highly discriminative features during inference.
When continually trained on a task sequence, FACET learns a single shared adapter that employs a dynamic task-conditioned feature transformation, shaping the overall feature distribution of the adapter into a mixture of overlap-reduced task-specific components.
On the other hand, we propose an efficient replay-free task-conditioned feature consistency loss, aiming to mitigate catastrophic forgetting of the learned mixture distribution in the adapter's feature space.
Even when maintaining only a single adapter, FACET demonstrates robust scalability. On both very long task sequences (e.g., 200 tasks) and standard short task sequences (e.g., 20 tasks), our method achieves superior performance while using significantly fewer trainable parameters and GFLOPs. The code will be made open source upon acceptance.
\end{abstract}

\section{Introduction}
In real-world applications, training data may arrive sequentially from continuously changing environments, which requires a learning system to acquire new concepts while retaining previous knowledge~\cite{lesort2020continualrobot,meng2025robotlifelongRL}. 
However, standard deep neural networks generally suffer from catastrophic forgetting: when trained on new data, their performance on previously learned tasks can severely degrade~\cite{li2017lwf,kirkpatrick2017EWC,rusu2016progressiveNN}. 
Continual learning (CL) aims to address this problem by enabling models to learn from continuous data streams without forgetting past knowledge. 
Among the many CL settings, class-incremental learning (CIL) is one of the most challenging because the model must learn new classes over time and, at test time, discriminate among all classes seen so far~\cite{masana2022CILsurvey}.

Traditional continual learning methods address catastrophic forgetting through regularization~\cite{kirkpatrick2017EWC,li2017lwf}, replay~\cite{rebuffi2017icarl,lopez2017gradientGEM},architectural expansion~\cite{rusu2016progressiveNN,yoon2017lifelongDEN}, and parameter isolation/sub-networks~\cite{aljundi2017expertGate,mallya2018packnet}.
More recently, pretrained-model (PTM) based approaches have become prevalent for CIL, as large pretrained backbones provide general-purpose, powerful representations, and require relatively few additional parameters when adapted to new tasks~\cite{wang2022L2P,smith2023coda,zhou2024EASE,zhou2025APER,wang2025tuna,sun2025mos,jiang2509MIN}.
A common strategy attaches task-specific learnable adapters to a frozen vision backbone~\cite{zhou2024EASE,wang2025tuna,jiang2509MIN,wang2025sema}, which produces explicit task-specific feature perturbations but requires a separate adapter per task, resulting in parameter and computation inefficiency.
Another popular solution trains separate LoRA modules~\cite{hu2022lora} by task and merges all learned modules before inference, exemplified by InfLoRA~\cite{liang2024inflora} and SD-LoRA~\cite{Wu2025SDLoRA}.
Although such methods improve parameter efficiency at inference time, an input is passed to the aggregated weights, which give rise to representation interference and diluted features that are suboptimal for classification.

To address these issues, we present \textbf{FACET}, a novel PTM-based CIL method that achieves parameter efficiency by using a single continually optimized adapter with an explicit task-conditioning mechanism. 
The core idea is to obtain task-dependent features by applying a dynamic task-conditioned transformation to the adapter's output. 
Specifically, we predict a task-dependent context vector for the input, and map it to a matrix that performs task-dependent channel-wise scaling and cross-channel mixing in the latent space of the adapter. 
This substantially improves the separability of task-specific feature distributions in the adapter's feature space, thereby reducing representation interference.

\begin{figure*}[t]
    \centering
    \includegraphics[width=0.99\textwidth]{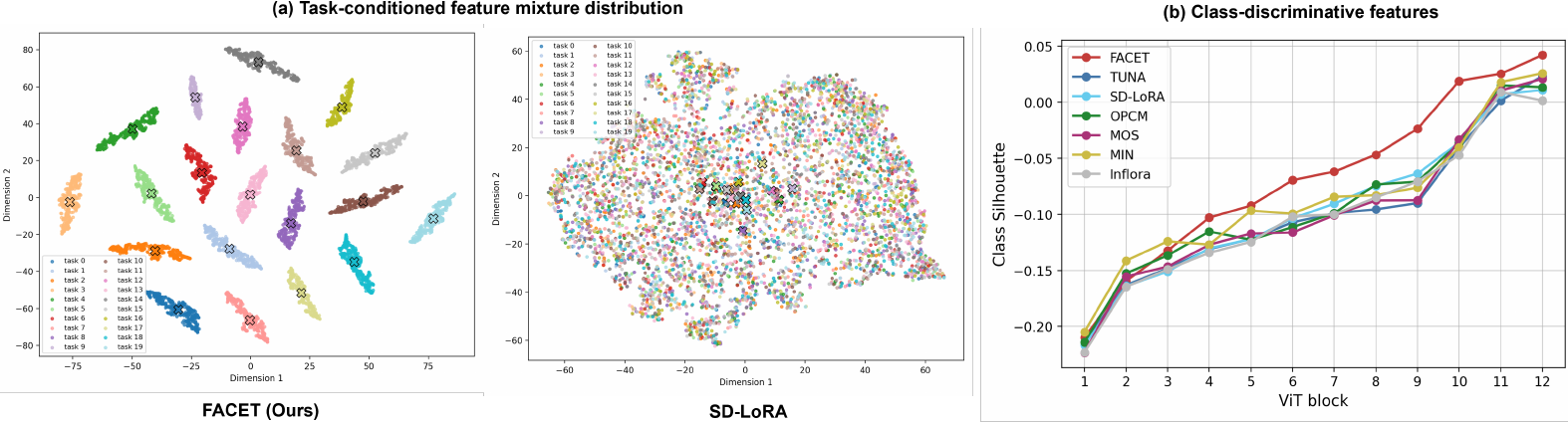}
     \vspace{-0.3em}
    \caption{Comparison of feature distributions. (a) t-SNE of block 1 features, colored based on \textbf{task}. FACET shows well-separated task-specific distributions, while SD-LoRA's are mixed.
    (b) \textbf{Class}-averaged silhouette score (higher values mean better intra‑class compactness and inter‑class separation) across blocks.
    FACET achieves higher scores than representative LoRA‑based (SD‑LoRA) and per‑task adapter based (TUNA) methods, producing more class‑discriminative features that are beneficial for classification.}
    \vspace{-1.5em}
\label{fig:intro_comparison}
\end{figure*}


However, training the shared adapter on a new task might distort the conditional feature distributions of previous tasks and make them less discriminative. To this end, the adapter is required to preserve the conditional feature distribution of any earlier task. That is, given any training sample from the new task, the adapter should be able to reproduce its feature representation as if conditioned on any earlier task. This goal is achieved via a simple \emph{task-conditioned feature consistency} objective.

We compare the representation quality of FACET with representative methods in Fig.~\ref{fig:intro_comparison}.
In early adapter blocks, FACET's task‑specific feature distributions are widely separated, while those of SD-LoRA are completely mixed (Fig.~\ref{fig:intro_comparison} (a)). %
As CIL ultimately requires discrimination among classes,  
Fig.~\ref{fig:intro_comparison} (b) compares the block-wise class-averaged Silhouette score~\footnote{The Silhouette score of a class is \((b-a)/\max(a,b)\), where \(a\) is the mean pairwise feature distance within the class and \(b\) is the mean pairwise distance between any feature in the class and any feature in the nearest different class.}. A higher score suggests more class‑discriminative features.
Driven by dynamic task conditioning, FACET achieves a significantly higher score in the last block, 
which indicates that features have transitioned from task‑scaffolded to class‑discriminative as depth increases. 

We evaluate FACET on standard class-incremental learning benchmarks.
FACET achieves the highest final accuracy on ImageNet-R, ImageNet-A, CIFAR-100, and ObjectNet.
Compared with SD-LoRA~\cite{Wu2025SDLoRA}, FACET improves the final accuracy from 75.26\% to 79.97\% on ImageNet-R with 20 tasks and from 55.96\% to 64.87\% on ImageNet-A.
Surprisingly, on OmniBenchmark-1K~\cite{lou2026CARE} with 200 tasks, FACET significantly improves over MIN~\cite{jiang2509MIN} and TUNA~\cite{wang2025tuna} by 3.45\% and 6.81\% in final accuracy, respectively. 
FACET is also $3.85\times$ faster and reduces GFLOPs by 62.69\% relative to MIN (Table~\ref{tab:task_efficiency_large}).
Note that MIN and TUNA learn a task-specific adapter for every task.

Our main contributions are as follows:

\noindent\textbf{A novel single‑adapter design with task‑conditioned transformations.}
We introduce a single, continually trained adapter that incorporates a dynamic task-conditioned transformation. This design achieves parameter efficiency and improves the separability of task-specific feature distributions, thus reducing representation interference.

\noindent\textbf{Conditioned feature consistency.}
We propose a conditioned feature consistency loss that aligns the feature representation of the latest adapter with its frozen snapshot under earlier task conditions, effectively mitigating forgetting.

\noindent\textbf{State‑of‑the‑art results.}
FACET achieves new state‑of‑the‑art experimental results on class‑incremental learning benchmarks with 10 to 200 tasks.

\section{Related Work}

\textbf{Pre-trained model-based Continual Learning.}
Early continual learning methods combat catastrophic forgetting through regularization~\cite{kirkpatrick2017EWC,zenke2017SI}, knowledge distillation~\cite{li2017lwf}, replay~\cite{rebuffi2017icarl,lopez2017gradientGEM}, parameter isolation/sub-networks~\cite{aljundi2017expertGate,mallya2018packnet,masse2018alleviatingSubnetwork,abati2020conditionalsubnetwork}, and architectural expansion~\cite{rusu2016progressiveNN,yoon2017lifelongDEN,yan2021DER}.

Based on the powerful generalization ability of pretrained models (PTMs), recent class‑incremental learning (CIL) methods attach a small number of learnable parameters for each task. For example, Prompt‑tuning approaches such as L2P~\cite{wang2022L2P}, DualPrompt~\cite{wang2022dualprompt}, and CODA-Prompt~\cite{smith2023coda} store a pool of task‑specific tokens and retrieve them at inference. Recent methods provide stronger learning capacity by incrementally appending task-specific adapters, i.e., EASE~\cite{zhou2024EASE} builds task‑specific subspaces by incrementally tuning adapters: TUNA~\cite{wang2025tuna}, MOS~\cite{sun2025mos}, and MIN~\cite{jiang2509MIN} assign a dedicated adapter to each task to prevent forgetting. 
However, these methods are parameter inefficient and computationally heavy as they must activate or aggregate all stored task-specific adapters at test time. 
In contrast, we use a single shared adapter with a dynamic task-conditioned feature transformation, producing separable task‑specific feature distributions without maintaining per‑task adapters.

\textbf{Parameter Efficient CIL.}
Some works strive for a single set of parameters in PTM‑based CIL. SLCA~\cite{zhang2023slca} and FeCAM~\cite{goswami2023fecam} focus on classifier alignment and covariance‑aware classifiers, respectively. 
Closer to our design, LoRA‑based merging methods (InfLoRA~\cite{liang2024inflora}, SD-LoRA~\cite{Wu2025SDLoRA}) and model merging techniques (MagMax~\cite{marczak2024magmax}, OPCM~\cite{Tang2025OPCM}) first train separate LoRA/adapters per task and then combine them into a single model during inference.
These methods apply a static merged model during inference, where the aggregated weights may cause representation interference and dilute the task-specificity of feature maps. 
In contrast, our method directly trains a single adapter for all tasks with explicit task conditioning, eliminating such interference without per‑task adapters.

\begin{figure*}[t]
    \centering
    \includegraphics[width=0.95\textwidth]{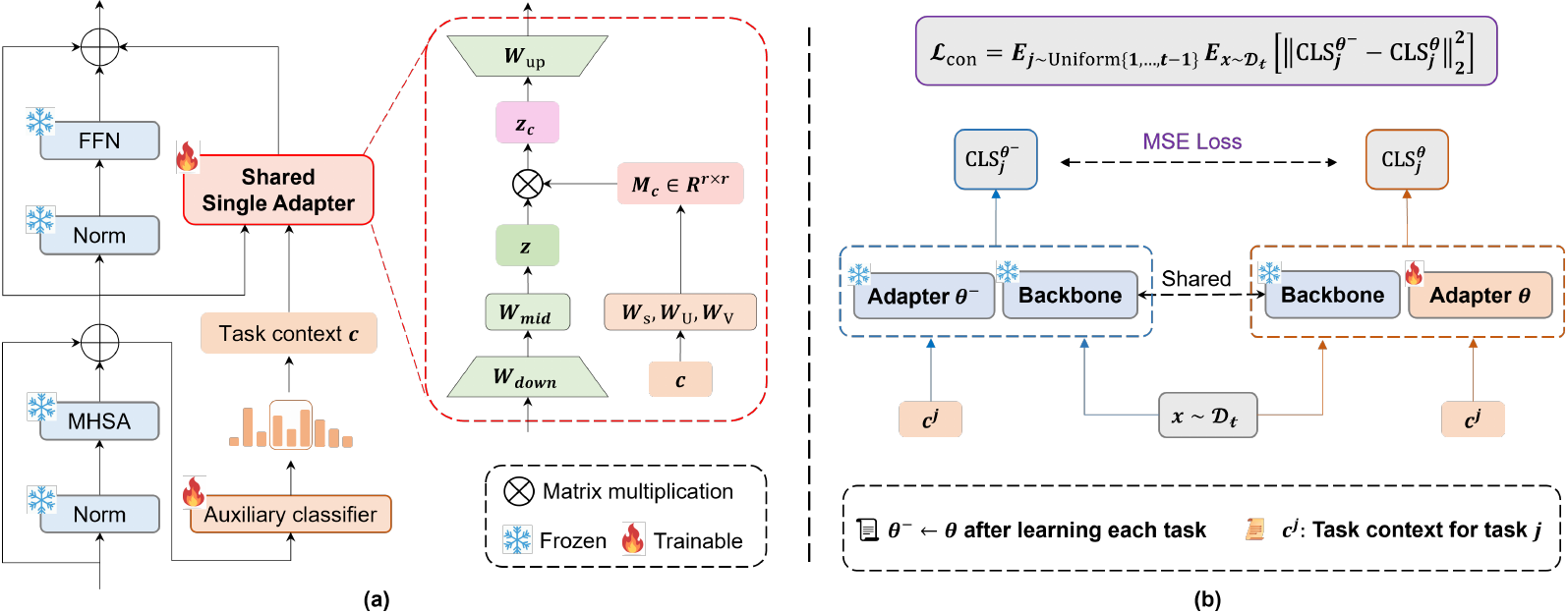}
    \caption{Illustration of FACET. \textbf{(a)} Task-conditioned feature transformation within a single adapter. We predict the task context $c$ and use it to generate a transformation matrix $M_c$ to obtain task-dependent features. Activation functions are not shown for simplicity. \textbf{(b)} Task-conditioned feature consistency objective when training on a new task $t$ with dataset $\mathcal{D}_t$.}
\label{fig:method_fig}
\vspace{-1.25em}
\end{figure*}

\section{Method}
\label{sec:method}

\subsection{Problem Formulation}
\label{sec:problem_formulation}

In class-incremental learning, a model learns from a data stream $\mathcal{D} = \{\mathcal{D}_1, \mathcal{D}_2, \ldots, \mathcal{D}_T\}$ consisting of $T$ tasks arriving sequentially.
Each task $t$ provides a training set $\mathcal{D}_t = \{(x_i^t,\, y_i^t)\}_{i=1}^{n_t}$, where $x_i^t$ is an image, $y_i^t \in \mathcal{Y}_t$ is its class label, and $n_t$ is the number of training samples.
The label sets of different tasks are disjoint: $\mathcal{Y}_s \cap \mathcal{Y}_t = \emptyset$ for all $s \neq t$.
We follow the exemplar-free setting~\cite{zhou2024EASE,wang2025tuna}. When task $t$ is learned, only the latest training data $\mathcal{D}_t$ are accessible.
We denote $\mathcal{Y}_{\mathrm{new}} = \mathcal{Y}_t$ as the set of new classes and $\mathcal{Y}_{\mathrm{old}} = \bigcup_{s=1}^{t-1} \mathcal{Y}_s$ as the set of previously seen classes.
After being trained on task $t$, the model is evaluated on all classes seen so far, $\mathcal{Y}_{\mathrm{old}} \cup \mathcal{Y}_{\mathrm{new}}$.

We decompose the model into a pretrained ViT backbone $\mathcal{F}$, an adapter with parameters $\theta$~\cite{chen2022adaptformer} inserted in parallel to the FFN in every transformer block, and a classifier $\mathbf{W}_t$ that grows by appending a new class vector for each task.
The backbone $\mathcal{F}$ is frozen throughout training. Only $\theta$ and $\mathbf{W}_t$ are updated.
The goal is to learn all $T$ tasks from the data stream and correctly classify any test sample into one of the classes in $\bigcup_{t=1}^T \mathcal{Y}_t$ without access to past data.


\subsection{Single Adapter with Dynamic Task-Conditioned Transformations}
\label{sec:task_conditioned_adapter}
As shown in Fig.~\ref{fig:method_fig} (a), a frozen pretrained Vision Transformer (ViT)~\cite{dosovitskiy2020vit} backbone \(\mathcal{F}\) is equipped with adapter modules placed in parallel to the feed‑forward network (FFN) in every transformer block. 
For an input \(x\) to the FFN, the output of the block \(o\) is
\begin{equation}
o = \operatorname{FFN}(x) + \operatorname{Adapter}(x, c) + x,
\label{eq:block_with_adapter}
\end{equation}
where \(c\) denotes a context vector that reflects the task-related information of $x$.
By conditioning the adapter's output explicitly on \(c\), we obtain task‑dependent features during inference while using only a \emph{single} shared set of adapter weights for all tasks, which is both parameter- and computation-efficient.

The hidden layer of a standard adapter~\cite{chen2022adaptformer} operates in a latent space, which often has much lower dimensionality than the ViT backbone.
To increase learning capacity without substantially increasing trainable parameters, we extend the standard adapter with one additional non‑linear hidden layer.
The revised adapter first computes a latent representation \(z \in \mathbb{R}^{r}\) as follows,
\begin{equation}
z = \sigma\bigl(\sigma(x W_{\text{down}})\, W_{\text{mid}}\bigr), \qquad
W_{\text{down}}\!\in\!\mathbb{R}^{d\times r},\; W_{\text{mid}}\!\in\!\mathbb{R}^{r\times r},
\label{eq:latent_z}
\end{equation}
where \(\sigma\) is the GELU~\cite{hendrycks2016GELU} activation, $d$ is the ViT's feature dimensionality, and \(r  < d\) is the latent dimensionality of the adapter. This is common computation for all tasks.
Next, before projecting the latent representation back to the ViT feature space, we apply a task-conditioned linear transformation to $z$,
\begin{equation}
z_c = zM_c,
\label{eq:task_condition_transformation}
\end{equation}
where $M_c \in \mathbb{R}^{r\times r}$ is a transformation matrix dynamically generated from the context vector $c$ and $z_c$ is the resulting task-dependent latent representation. 
Context vectors related to different tasks give rise to different matrix transformations. As a result, the same latent representation $z$ can be transformed to different task-dependent latent features. The adapter output is obtained by projecting $z_c$ up to the ViT feature space through the shared up-projection $W_{up}\in \mathbb{R}^{r\times d}$:
\begin{equation}
    \text{Adapter}(x,c)  = (z_c)W_{up}.
    \label{eq:adapter_output}
\end{equation}
Since the up-projection is linear, applying the task-conditioned transformation in the latent space is sufficient to make the final adapter output task-dependent, while all adapter weights remain shared across tasks.

Let us now discuss more details about context vectors and task-conditioned transformations.
The context vector reflects information on the task identity of the given input. A simple option is to set it to be the task identity itself, i.e., \(c \in \{0,1\}^{K_{\max}}\), where \(K_{\max}\) is a sufficiently large upper bound on the total number of tasks (ablation of \(K_{\max}\) is given in Appendix Table \ref{tab:abl_kmax}).
During training, the task identity is known by counting the number of trained tasks at present, while lightweight task-wise classifiers predict the context vector during inference (Sec.~\ref{sec:two_pass_inference}).

Given a context vector, we construct the task-conditioned transformation \(M_c\) in a low‑rank decomposition form to maintain parameter efficiency.
\begin{equation}
M_c = \operatorname{Diag}(\mathbf{s}) + U V^{\top},
\label{eq:M_c}
\end{equation}
where \(\mathbf{s} \in \mathbb{R}^{r}\) determines task-conditioned scaling, which independently rescales each feature dimension in a task-dependent manner, and \(U, V \in \mathbb{R}^{r \times k}\) with \(k \ll r\) implement cross‑dimension mixing, which redistributes activations across dimensions so that different tasks produce different linear combinations of the latent feature dimensions. 
The above matrix components are directly mapped from the context vector via linear projections,
\[
\mathbf{s} = W_s c, \qquad U = \operatorname{reshape}(W_U c), \qquad V = \operatorname{reshape}(W_V c),
\]
where $ W_s \in \mathbb{R}^{r \times K_{\max}}$ and $W_U,W_V \in \mathbb{R}^{r k \times K_{\max}}$ are learnable projection matrices, and \(\operatorname{reshape}(\cdot)\) converts a vector of length \(r k\) into a \(r \times k\) matrix.
Thus, a single shared adapter, without any task-specific weights, produces task‑conditioned feature transformations and, in turn, task-dependent final representations entirely on the basis of the context vector. 
This is the core difference between our method and other LoRA-based merging methods.

To improve training stability, we initialize \(W_s, W_U, W_V\) to zero so that the adapter output is zero during the first training step. We also add small Gaussian noise to the context vector during training, aiming to improve robustness against imperfect task identity predictions during inference and softly enable backward knowledge transfer during training. 

\paragraph{Remark} 
As a single adapter is shared among all learned tasks and each task has its own feature distribution, the overall distribution of the adapter's final output features can be viewed as a mixture of task-specific components. Task-conditioned feature transformations are capable of reducing overlap among these task-specific components by altering their location and shape.
Our experiments confirm that introducing task-conditioned transformations better separates task-specific components (Fig.~\ref{fig:intro_comparison}).

We also considered adding explicit losses to improve the separability of the mixture components, such as maximizing their cosine distance or maximizing the orthogonality of task-conditioned transformations. Nevertheless, this reduces the learning capacity of the adapter and leads to sub-optimal results, as shown in our ablation studies (Table~\ref{tab:abl_dcom_form}).

Using context signals that reflect task identity as conditioning has been previously studied in continual learning. For instance, using them as learnable context tokens concatenated to the input of the model~\cite{wang2022L2P,wang2022dualprompt,smith2023coda}, as a gating mechanism to select subnetworks~\cite{aljundi2017expertGate,abati2020conditionalsubnetwork,masse2018alleviatingSubnetwork}, or to route inputs to task‑specific adapters~\cite{wang2025tuna,sun2025mos}. Distinct from previous works, FACET applies context signals to transform the adapter's latent representation, making task-specific feature distributions more separable, which is a direction that has not been explored.

\subsection{Task-Conditioned Feature Consistency}
\label{sec:replay_free_distillation}

Since the adapter weights are tuned on every new task, there is an increasing risk of gradually shifting the adapter's latent feature distribution. 
This may lead to forgetting of the learned mixture distribution because the task-specific mixture components are dependent on the adapter weights. 
To this end, we introduce a simple \emph{task‑conditioned feature consistency} objective that preserves the learned task-specific mixture components without resorting to any historical data.
This allows our adapter to produce reliable task-dependent features for continual learning.
Specifically, after learning task \(t-1\), we make a copy of the latest adapter weights, \(\theta^- \leftarrow \theta\), and keep the copy frozen when tuning the adapter on task \(t\).  
For each mini‑batch, we uniformly sample the index of a previous task \(j \sim \mathrm{Uniform}\{1,\ldots,t-1\}\) and construct the corresponding context vector \(c^j \in \mathbb{R}^{K_{max}}\). 
Then the training samples for task \(t\) pass through the following two paths:

\textbf{Main path}: a training sample $x \in \mathcal{D}_t$ is passed to the ViT backbone equipped with the adapter (\(\theta\)) currently being trained on task \(t\) and use \(c^j\) as the context vector. The [CLS] token from the last transformer block is the main feature, and is denoted as $\text{CLS}_j^{\theta}$ 

\textbf{Reference path}: $x$ is passed to the ViT backbone equipped with the frozen adapter snapshot (\(\theta^-\)) and use \(c^j\) as the context vector again. The [CLS] token from the last transformer block is taken as the reference feature, and is denoted as $\text{CLS}_j^{\theta^-}$ 

Evaluating both paths on the same input under the same task conditioning enables us to directly measure how much the latest adapter has drifted from the learned feature distribution of task \(j\).  
This does not require \(x\) to be a sample from task \(j\) as the context vector \(c^j\) drives the adapter to process \(x\) as a sample from task \(j\) regardless of its true task identity.
Our experiments demonstrate that this simple strategy can preserve the feature distribution of historical tasks without using their original data (Appendix~\ref{app:snapshot}) and mitigates forgetting substantially (Table~\ref{tab:abl_components}).

The training loss is the squared Euclidean distance between conditioned [CLS] tokens averaged uniformly over all previous tasks:
\begin{equation}
\mathcal{L}_{\mathrm{con}} = \mathbb{E}_{j \sim \mathrm{Uniform}\{1,\ldots,t-1\}}\,\mathbb{E}_{x \sim \mathcal{D}_t}\!\left[\left\| \text{CLS}_j^{\theta^-} - \text{CLS}_j^{\theta} \right\|_2^2\right].
\label{eq:distill_loss}
\end{equation}
In practice, each mini‑batch draws a single pair \((j, x)\), yielding an unbiased stochastic estimate of this expectation and providing uniform coverage of all earlier tasks over the course of training.  
We sample only one previous task per iteration rather than considering all previous tasks simultaneously because the latter would over‑constrain \(\theta\) and reduce the adapter's capacity to learn the latest task.


The overall training loss is 
\begin{equation}
\mathcal{L} = \mathcal{L}_{\text{cls}} + \lambda_{\text{con}}\,\mathcal{L}_{\text{con}},
\label{eq:final_loss}
\end{equation}
where \(\mathcal{L}_{\text{cls}}\) is a weighted cross‑entropy loss combined with classifier alignment as in~\cite{zhang2023slca,wang2025tuna,sun2025mos} (applied to learn a new conditional feature distribution for task $t$), and \(\lambda_{\text{con}}=2\) in our experiments.

Note that our task-conditioned feature consistency differs from standard distillation methods such as LwF~\cite{li2017lwf}, which aligns global predictions or features~\cite{heo2019comprehensiveFeatDistill,li2024continualKD_survey}, but cannot preserve the \textit{task‑specific feature distributions} that are central to our design.

\subsection{Oracle-Free Inference}
\label{sec:two_pass_inference}
Generating task-dependent features from FACET at each block requires a task context vector \(c\). We therefore introduce a straightforward strategy to estimate \(c\) during inference. 
Specifically, we attach a linear classifier to the output of the attention layer within each transformer block (Fig.~\ref{fig:method_fig}). 
This auxiliary classifier receives the same supervision alongside the final classification head during the continual training phase. 
During inference, this classifier outputs a probability distribution $p_t$ for all classes $\mathcal{Y}_t$ for all learned tasks $t \in \{1,...,T\}$.
We compute the entropy \(H_t=-\sum_{y\in\mathcal{Y}_t}p_t(y)\log p_t(y)\) for each task and select the task context with the highest prediction certainty (lowest entropy) \(\hat{t}=\arg\min_t H_t\).
The final classifier predicts from the last block's [CLS] feature.
Notably, the auxiliary classifiers contain task-specific parameters with a total of $(d\times \sum_t{|\mathcal{Y}_t|})$ parameters, where $\sum_t{|\mathcal{Y}_t|}$ is the total number of classes in the dataset. Table~\ref{tab:task_efficiency_large} shows that FACET significantly reduces the task-specific costs compared to existing task-specific adapter methods.
Section~\ref{sec:ablation_studies} compares alternative context-prediction strategies.

\section{Experiments}
\label{sec:experiments}

We evaluate FACET on standard CIL benchmarks covering short to very long task sequences.
Despite employing a single adapter, FACET outperforms strong baselines, especially on long task sequences.
Ablation studies validate the effectiveness of all components.

\subsection{Experimental Setting}
\label{sec:exp_setting}

\textbf{Datasets.}
We benchmark on four widely used datasets~\cite{wang2022L2P,smith2023coda,zhou2024EASE,liang2024inflora,jiang2509MIN}, including CIFAR-100~\cite{krizhevsky2009CIFAR100}, ImageNet-R~\cite{hendrycks2021ImagenetR}, ImageNet-A~\cite{hendrycks2021ImagenetA}, and ObjectNet~\cite{barbu2019objectnet}, plus long task sequences with up to 200 tasks in a recent benchmark, OmniBenchmark-1K~\cite{lou2026CARE} .
Together, these benchmarks cover short, medium-length, and long task sequences in standard and distribution-shifted visual domains.

\textbf{Baselines.}
We compare FACET against state-of-the-art prompt-based and adapter-based PTM CIL methods with task-specific parameters (L2P~\cite{wang2022L2P}, DualPrompt~\cite{wang2022dualprompt}, CODA-Prompt~\cite{smith2023coda}, EASE~\cite{zhou2024EASE}, APER~\cite{zhou2025APER}, TUNA~\cite{wang2025tuna}, MOS~\cite{sun2025mos}, and MIN~\cite{jiang2509MIN}), classifier-based methods (SLCA~\cite{zhang2023slca}, FeCAM~\cite{goswami2023fecam}), and LoRA-based weight-merging methods (InfLoRA~\cite{liang2024inflora}, SD-LoRA~\cite{Wu2025SDLoRA}, MagMax~\cite{marczak2024magmax}, and OPCM~\cite{Tang2025OPCM}).

\textbf{Implementation Details.}

We run all our experiments on a single NVIDIA L40S GPU.
Following prior work~\cite{zhou2024EASE,zhou2025APER,wang2025tuna,jiang2509MIN,sun2025mos}, we employ the representative pre-trained vision backbone ViT-B/16-IN21K~\cite{dosovitskiy2020vit}.
FACET adds a single adapter per transformer block (latent dimension 384, hidden depth 2) with $k=8$ and $K_{\max}=200$ for the task-conditioned transformation matrix $M_c$. 
For training, we use SGD (momentum 0.9, weight decay \(5\times10^{-4}\)), cosine learning‑rate decay with base learning rate 0.02 and batch size 16, and conditional feature consistency weight \(\lambda_{\text{con}}=2.0\).
The classification head follows~\cite{zhou2024EASE,wang2025tuna}.
More details about datasets and settings are in the Appendix.

\textbf{Evaluation Protocol and Performance Metrics.}
We follow standard settings~\cite{wang2022L2P,wang2025tuna,jiang2509MIN} to partition a dataset into $T$ tasks with the learning order determined by the random seed 1993. 
We denote Inc $n$ as each task having $n$ classes.  
For FACET, we report the average performance over 3 runs. 
We report the average incremental accuracy $\bar{A}=\frac{1}{T}\sum_{t=1}^{T} A_t$ and final accuracy $A_T$, where $A_t$ is the average accuracy over all classes seen so far after the adapter has been trained on task $t$.
Due to space limit, results on the ViT-B/16-IN1k backbone and other random seeds are provided in the Appendix~\ref{appendix_robust}.

\begin{table*}[t]
  \centering
  \caption{Performance comparison with state-of-the-art methods on standard benchmarks.}
  \resizebox{0.985\textwidth}{!}{
    \begin{tabular}{l|cccc|cccc|cc|cc}
    \toprule
    \multirow{3}[2]{*}{Method} & \multicolumn{4}{c|}{\textbf{ImageNet-R}} & \multicolumn{4}{c|}{\textbf{CIFAR-100}} & \multicolumn{2}{c|}{\textbf{ImageNet-A}} & \multicolumn{2}{c}{\textbf{ObjectNet}} \\
          & \multicolumn{2}{c}{\textbf{10 Tasks (Inc20)}} & \multicolumn{2}{c|}{\textbf{20 Tasks (Inc10)}} & \multicolumn{2}{c}{\textbf{10 Tasks (Inc10)}} & \multicolumn{2}{c|}{\textbf{20 Tasks (Inc5)}} & \multicolumn{2}{c|}{\textbf{10 Tasks (Inc20)}} & \multicolumn{2}{c}{\textbf{10 Tasks (Inc20)}}  \\
          & $\mathbf{\bar{A}}$      & $\mathbf{{A}}_T$    & $\mathbf{\bar{A}}$      & $\mathbf{{A}}_T$    & $\mathbf{\bar{A}}$      & $\mathbf{{A}}_T$    & $\mathbf{\bar{A}}$      & $\mathbf{{A}}_T$    & $\mathbf{\bar{A}}$      & $\mathbf{{A}}_T$    & $\mathbf{\bar{A}}$      & $\mathbf{{A}}_T$  \\
    \midrule
    L2P\textsubscript{(CVPR'22)} & 75.46 & 69.77 & 63.75 & 55.78 & 85.92 & 79.19 & 85.94 & 79.93 & 49.39 & 41.71 & 66.77 & 55.16 \\
    DualPrompt\textsubscript{(ECCV'22)} & 73.10 & 67.18 & 66.52 & 61.77 & 89.65 & 84.89 & 87.87 & 81.15 & 53.71 & 41.67 & 64.31 & 52.99 \\
    CODA-Prompt\textsubscript{(CVPR'23)} & 77.97 & 72.27 & 70.45 & 64.68 & 91.05 & 86.44 & 89.11 & 81.96 & 53.54 & 42.73 & 66.53 & 56.80 \\
    EASE\textsubscript{(CVPR'24)} & 81.74 & 76.17 & 81.18 & 74.62 & 92.11 & 87.72 & 91.51 & 85.80 & 65.34 & 55.04 & 71.04 & 59.37 \\
    SEMA\textsubscript{(CVPR'25)} & 81.39 & 77.84 & 77.84 & 69.60 & 91.60 & 86.75 & 92.23 & 87.84 & 63.83 & 52.21 & 67.95 & 54.92 \\
    MOS\textsubscript{(AAAI'25)} & 81.74 & 76.17 & 82.96 & 77.93 & 93.83 & 90.19 & 93.30 & 89.25 & 69.13 & 59.12 & 74.75 & 65.10 \\
    TUNA\textsubscript{(ICCV'25)} & 84.22 & 79.42 & 82.54 & 77.45 & 94.85 & 91.71 & 94.44 & 90.74 & 72.96 & 64.38 & 76.46 & 66.32 \\
    MIN\textsubscript{(NeurIPS'25)} & 85.18 & 79.75 & 83.69 & 78.08 & 95.12 & 92.12 & 94.31 & 91.03 & 72.89 & 64.32 & 72.56 & 61.36 \\
    SLCA\textsubscript{(ICCV'23)} & 81.17 & 77.00 & 81.85 & 76.63 & 92.67 & 89.30 & 93.52 & 90.07 & 68.66 & 58.74 & 74.12 & 63.23 \\
    FeCAM\textsubscript{(NeurIPS'23)} & 79.02 & 72.53 & 77.84 & 71.05 & 93.23 & 89.05 & 91.86 & 87.04 & 56.04 & 46.41 & 68.68 & 57.39 \\
    APER-Adapter\textsubscript{(IJCV'25)} & 75.82 & 67.95 & 72.35 & 64.33 & 92.22 & 87.45 & 90.65 & 85.15 & 60.53 & 49.57 & 69.24 & 57.41 \\
    MagMax-LoRA\textsubscript{(ECCV'24)} & 84.59 & 77.68 & 81.8 & 73.40 & 94.76 & 91.41 & 94.31 & 90.94 & 66.64 & 54.64 & 70.85 & 59.15 \\
    OPCM-LoRA\textsubscript{(NeurIPS'25)} & 84.37 & 79.38 & 84.08 & 77.62 & 94.90 & 91.92 & 94.60 & 91.18 & 69.67 & 59.78 & 71.99 & 60.28 \\
    InfLoRA\textsubscript{(CVPR'24)} & 80.82 & 75.65 & 77.28 & 71.01 & 91.70 & 86.51 & 89.13 & 81.46 & 58.50 & 46.28 & 70.67 & 58.04 \\
    SD-LoRA\textsubscript{(ICLR'25)} & 82.04 & 77.34 & 80.22 & 75.26 & 92.54 & 88.01 & 90.90 & 85.18 & 64.95 & 55.96 & 70.37 & 58.54 \\
    \rowcolor[rgb]{ 0.8, 0.9, 0.95}\textbf{FACET (Ours)} & \textbf{85.41} & \textbf{80.66} & \textbf{84.73} & \textbf{79.97} & \textbf{95.18} & \textbf{92.27} & \textbf{94.62} & \textbf{91.31} & \textbf{73.31} & \textbf{64.87} & \textbf{77.13} & \textbf{67.38} \\
    \bottomrule
    \end{tabular}
    }
  \label{tab:standard_benchmark}%
  \vspace{-0.75em}
\end{table*}

\subsection{Main Results}

Tables~\ref{tab:standard_benchmark} and~\ref{tab:long_eval_2} compare FACET with SOTA PTM-based CIL methods. Our experimental results clearly demonstrate two key advantages. First, our single adapter yields superior performance on standard benchmarks in comparison to computationally heavy methods based on a set of task-specific adapters. 
Second, this design surprisingly achieves leading performance on long task sequences of up to 200 tasks when compared to other strong baselines. More details are provided below.

\begin{table*}[t]
  \centering
  \caption{Performance comparison on benchmarks with long task sequences. InfLoRA and SD-LoRA failed to converge on the 200 task setting.}
  \scalebox{0.7}{
    \begin{tabular}{l|cc|cc|cc|cc}
    \toprule
    \multirow{3}[2]{*}{Method}
      & \multicolumn{2}{c|}{\textbf{ImageNet-R}}
      & \multicolumn{2}{c|}{\textbf{ImageNet-A}}
      & \multicolumn{4}{c}{\textbf{OmniBenchmark-1K}} \\
      & \multicolumn{2}{c|}{\textbf{50 Tasks (Inc4)}}
      & \multicolumn{2}{c|}{\textbf{50 Tasks (Inc4)}}
      & \multicolumn{2}{c|}{\textbf{100 Tasks (Inc10)}}
      & \multicolumn{2}{c}{\textbf{200 Tasks (Inc5)}} \\
      & $\mathbf{\bar{A}}$ & $\mathbf{A}_T$
      & $\mathbf{\bar{A}}$ & $\mathbf{A}_T$
      & $\mathbf{\bar{A}}$ & $\mathbf{A}_T$
      & $\mathbf{\bar{A}}$ & $\mathbf{A}_T$ \\
    \midrule
    L2P\textsubscript{(CVPR'22)}          & 69.16 & 63.45 & 49.89 & 36.41 & 60.91 & 48.87 & 57.53 & 45.25 \\
    DualPrompt\textsubscript{(ECCV'22)}   & 64.00 & 56.33 & 43.85 & 29.95 & 62.18 & 49.45 & 59.13 & 45.62 \\
    CODA-Prompt\textsubscript{(CVPR'23)}  & 62.43 & 57.57 & 38.24 & 26.60 & 64.16 & 51.75 & 60.22 & 47.56 \\
    EASE\textsubscript{(CVPR'24)}         & 78.11 & 70.63 & 59.86 & 47.53 & 65.00 & 53.54 & 67.23 & 55.13 \\
    SEMA\textsubscript{(CVPR'25)}         & 67.80 & 59.32 & 52.99 & 40.68 & 56.55 & 33.96 & 45.06 & 19.95 \\
    MOS\textsubscript{(AAAI'25)}          & 75.15 & 66.80 & 63.72 & 51.22  & 76.80 & 64.27 & 75.92 & 63.51 \\
        TUNA\textsubscript{(ICCV'25)}         & 80.93 & 75.48 & 67.82 & 58.72 & 71.93 & 60.04 & 71.45 & 59.14 \\
    MIN\textsubscript{(NeurIPS'25)}          & 82.26 & 75.72 & 66.15 & 57.08  & 75.46 & 63.60 & 74.94 & 62.50 \\
    APER-Adapter\textsubscript{(IJCV'25)} & 72.43 & 64.83 & 61.29 & 48.58  & 73.23 & 62.24 & 72.49 & 61.53 \\
    MagMax-LoRA\textsubscript{(ECCV'24)}  & 81.87 & 74.97 & 65.40 & 52.58  & 69.93 & 41.85 & 61.30 & 30.27 \\
    OPCM-LoRA\textsubscript{(NeurIPS'25)}  & 82.49 & 73.70  & 66.58 & 54.02 & 60.19 & 33.73 & 58.32 & 28.43 \\
    InfLoRa\textsubscript{(CVPR'24)}  & 71.68 & 62.62 & 50.17 & 38.70 & 51.53 & 27.01 & - & - \\
    SD-LoRA\textsubscript{(ICLR'25)}  & 68.40 & 63.28 &  54.72 & 41.28 & 53.97 & 28.15 & - & - \\
    
    
    \rowcolor[rgb]{0.8, 0.9, 0.95}\textbf{FACET (Ours)} & \textbf{82.70} & \textbf{76.63} & \textbf{69.16} & \textbf{59.58} & \textbf{76.88} & \textbf{66.17} & \textbf{77.30} & \textbf{65.95} \\
    \midrule
    \bottomrule
    \end{tabular}%
  }
  \label{tab:long_eval_2}%
  \vspace{-0.75em}
\end{table*}

\textbf{Comparison with State-of-the-Art Methods.}
Table~\ref{tab:standard_benchmark} shows that FACET achieves the highest final accuracy $A_T$ on all four datasets that have 10 to 20 tasks.
For instance, on ImageNet-R with 20 tasks, FACET improves $A_T$ from 78.08\% to 79.97\% compared to MIN, one of the strongest baselines that stores one adapter-like module for every task.
On the more challenging ImageNet-A benchmark, where adversarially filtered images make class discrimination more difficult, the strongest baselines with task-specific adapters, TUNA and MIN, substantially outperform LoRA-based merging methods, showing that explicit task-specific representations are important.
Nevertheless, FACET achieves the highest $A_T$ (64.87\%) and competitive $\bar{A}$ (73.31\%) with a single adapter, improving over the strongest LoRA-based merging method by 3.64\% in $\bar{A}$ and 5.09\% in $A_T$.
OPCM-LoRA remains competitive on ImageNet-R and CIFAR-100, but FACET still improves $A_T$ by 2.35\% on ImageNet-R with 20 tasks.

\textbf{Scaling to Long Task Sequences.}
Table~\ref{tab:long_eval_2} evaluates whether FACET remains effective as the number of tasks in a sequence increases to 100 or even 200, when storing, searching over, or merging many task-specific modules becomes increasingly difficult.
On ImageNet-R with 50 tasks, FACET achieves the highest $\bar{A}$ and $A_T$. It improves $A_T$ from 75.72\% to 76.63\% compared to MIN, and by an absolute 2.93\% compared to OPCM-LoRA, the strongest LoRA-based merging method.
On ImageNet-A with 50 tasks, FACET again achieves the highest $\bar{A}$ and $A_T$, improving over TUNA by 1.34\% in $\bar{A}$ and 0.86\% in $A_T$.
This is significant because ImageNet-A is the setting where LoRA-based merging methods struggle the most: FACET improves $A_T$ over OPCM-LoRA by 5.56\% while preserving the single-adapter design.
The scaling advantage becomes clearer for 100 and 200 tasks.
On OmniBenchmark-1K with 100 tasks, FACET improves over the strongest baseline by 1.90\% in $A_T$, while achieving the highest $\bar{A}$ of 76.88.
When the number of tasks further increases to 200, FACET remains the top performer among all baselines, improving $\bar{A}$ from 75.92\% to 77.30\% and $A_T$ from 63.51\% to 65.95\% over MOS.
These results show that FACET is capable of scaling to a large number of tasks while maintaining a single adapter, and task-conditioned feature transformations remain effective even when the number of tasks reaches 200.

\begin{table}[t]
\centering
\scriptsize
\setlength{\tabcolsep}{3.5pt}
\caption{Comparison of parameter, computation, and GFLOPs among SOTA methods.
Latency is the median latency at batch size 32.}
\label{tab:task_efficiency_large}

\begin{minipage}[t]{0.48\columnwidth}
\centering
\textit{ImageNet-R (50 tasks)}\\[2pt]
\begin{tabular}{lcccccc}
\toprule
Method & Params(M) & Latency(ms) & GFLOPs & $\bar{A}$ & $A_T$ \\
\midrule
\rowcolor[rgb]{0.8, 0.9, 0.95} Ours
& \textbf{10.20} & \textbf{28.61} & \textbf{37.37} & \textbf{82.70} & \textbf{76.63} \\
MoS
& 15.37 & 1431.30 & 1798.57 & 75.15 & 66.80 \\
TUNA
& 15.37 & 1308.28 & 1798.05 & 80.93 & 75.48 \\
MiN
& 135.15 & 44.02 & 55.39 & 82.26 & 75.72 \\
\bottomrule
\end{tabular}
\end{minipage}
\hfill
\begin{minipage}[t]{0.48\columnwidth}
\centering
\textit{OmniBenchmark-1K (200 tasks)}\\[2pt]
\begin{tabular}{lcccccc}
\toprule
Method & Params(M) & Latency(ms) & GFLOPs & $\bar{A}$ & $A_T$ \\
\midrule
\rowcolor[rgb]{0.8, 0.9, 0.95} Ours
& \textbf{22.88} & \textbf{36.64} & \textbf{40.23} & \textbf{77.30} & \textbf{65.95} \\
MoS
& 61.63 & 5717.77 & 7088.81 & 75.92 & 63.51 \\
TUNA
& 61.63 & 6306.79 & 7086.94 & 71.45 & 59.14 \\
MiN
& 1831.59 & 141.42 & 107.84 & 74.94 & 62.50 \\
\bottomrule
\end{tabular}
\end{minipage}
\vspace{-1.5em}
\end{table}

\subsection{Parameter, Computation and GFLOPs}
\label{sec:efficiency}
Table~\ref{tab:task_efficiency_large} compares our method with the best-performing baselines (i.e., MIN, MOS, and TUNA) in trainable parameter count, inference latency, and GFLOPs per image on long task sequences. 
Our method is among the most efficient and achieves a $156\times$ speedup in inference latency and only use $0.57$\% GFLOPs compared to MoS when there are 200 tasks. 
Furthermore, FACET requires only 1.25\% of the cumulative trainable parameters of MIN. 
These findings demonstrate that our proposed method maintains leading performance while delivering better parameter efficiency and substantially lower latency than per-task adapter methods such as TUNA/MOS.

\subsection{Ablation Studies}
\label{sec:ablation_studies}
We extensively verify the effectiveness of all components in FACET with ablation studies conducted on the ImageNet-R Inc10 (20 tasks) dataset, which provides a challenging setting.
\begin{wraptable}{r}{0.73\columnwidth}
  \centering
  \vspace{-0.9em}
  \caption{Incremental component ablation on ImageNet-R (Inc10).}
     \scalebox{0.825}{
    \begin{tabular}{lccc|cc}
    \toprule
    Variant & $M_c$ cond & Consistency & Other designs & $\mathbf{\bar{A}}$ & $\mathbf{A}_T$ \\
    \midrule
    Plain Single Adapter              & $\times$     & $\times$  & $\times$  & 57.75 & 48.75 \\
    +\,$M_c$ Transformation            & \checkmark   & $\times$ & $\times$   & 70.06 & 60.02 \\
    +\,Cond. Feat. Consis.                   & \checkmark & \checkmark & $\times$ & 83.90 & 78.23 \\
    \rowcolor[rgb]{0.8, 0.9, 0.95}\textbf{FACET (Full)} &  \checkmark & \checkmark  & \checkmark & 84.70 & 79.97 \\
    Feat. Consis. Only       & $\times$   & \checkmark   & \checkmark & 72.94 & 65.39 \\
    \bottomrule
    \end{tabular}%
    }
    \vspace{-1.5em}
  \label{tab:abl_components}
\end{wraptable}
The baseline is our full FACET method presented in Table~\ref{tab:standard_benchmark}.

\paragraph{Individual contributions of components.}
Table~\ref{tab:abl_components} decomposes the gains of FACET. A plain single adapter reaches only 48.75\% final accuracy, confirming that simply sharing weights across tasks is insufficient.
Adding task-conditioned feature transformations ($M_c$ transformation) improves final accuracy to 60.02\%, demonstrating that explicit task conditioning and conditioned transformations yield a substantial benefit by making task-specific feature distributions more separable.
The performance jump to 78.23\% after adding task-conditioned feature consistency, showing that once FACET creates separable mixture components, they are fragile to forgetting and must be explicitly preserved. 
Note that non-conditioned feature consistency alone achieves an $A_T$ of 65.39\%, indicating that the interplay between conditioned feature transformation and consistency is crucial.
Finally, other designs such as adding gaussian noise during training together contributes a further improvement to 79.97\%.

\paragraph{Variants of conditioned feature transformation.}
Table~\ref{tab:abl_dcom_form} studies alternative forms of task-conditioned transformation and the effect of an explicit separability loss (Sec.~\ref{sec:task_conditioned_adapter}), reporting $\bar{A}$, $A_T$, and the silhouette scores computed from last-block features as a measure of class discriminability.
\begin{wraptable}{r}{0.55\columnwidth}
  \centering
  \vspace{-0.75em}
  \caption{Ablation on conditioned feature transformation and separability loss.}
   \scalebox{0.825}{
    \begin{tabular}{lccc}
    \toprule
    Form & $\mathbf{\bar{A}}$ & $\mathbf{A}_T$ & Silhouette\\
    \midrule
    No transformation & 72.49 & 65.39& -0.11\\
    Concat $c$ to adapter input & 69.46 & 66.42 & -0.086\\
    $M_c$ at adapter input  & 72.27 & 69.22 & -0.057 \\
    $M_c=UV^\top$                          & 84.50 & 78.58 & 0.026\\
    $M_c =\mathrm{Diag}(s)$               & 84.46 & 79.60 & 0.040 \\
    \rowcolor[rgb]{0.8, 0.9, 0.95}$M_c =\mathrm{Diag}(s){+}UV^\top$ & 84.70 & 79.97 & 0.042 \\
    \midrule
    $+$Orthogonal $M_c$ loss &82.19 & 78.63 & 0.035\\
    $+z_c$ cos distance loss & 81.83 & 78.02 & 0.014\\
    \bottomrule
    \end{tabular}%
    }
  \label{tab:abl_dcom_form}
  \vspace{-0.75em}
\end{wraptable}
Transforming the adapter input instead of its output causes a sharp drop of $A_T$ to 69.22\% and a negative silhouette score, suggesting that intra-class distances exceed inter-class distances.
This confirms that conditioning the adapter output via Eq.~\eqref{eq:task_condition_transformation} is essential.
Diagonal-only and low-rank-only variants of $M_c$, already raise performance clearly above the plain single adapter, showing that the primary gain comes from explicit task conditioning.
Finally, adding explicit losses to further separate task-specific feature distributions hurts performance since it reduces the learning plasticity for new tasks.

\begin{wraptable}{r}{0.4\columnwidth}
  \centering
  \vspace{-0.75em}
  \caption{Ablation on conditioned feature consistency.}
  \scalebox{0.825}{
  \begin{tabular}{lcc}
    \toprule
    Strategy & $\mathbf{\bar{A}}$ & $\mathbf{A}_T$ \\
    \midrule
    Most recent ($t-1$)    & 83.31 & 77.03 \\
    All tasks simultaneous & 84.01 & 78.37 \\
    Uniform, EMA teacher   & 83.96 & 78.85 \\
    \rowcolor[rgb]{0.8, 0.9, 0.95}Uniform, snapshot (ours) & 84.70 & 79.97 \\
    \midrule
    All features (Last Block) & 80.56 & 70.70 \\
    CLS in every block & 84.35 & 79.80 \\
    \midrule
    Cosine Distance & 84.53 & 79.46 \\
    \bottomrule
  \end{tabular}
  }
  \label{tab:abl_distill_sampling}
\end{wraptable}

\paragraph{Design of conditioned feature consistency.}
Table~\ref{tab:abl_distill_sampling} studies how to best preserve previously learned task-specific feature distributions.
Only enforcing consistency for the most recent task under-protects earlier task-specific distributions ($A_T=77.03\%$), while considering all previous tasks simultaneously over-constrains the adapter and limits learning plasticity ($A_T=78.37\%$).
Uniformly sampling one previous task per mini-batch achieves the best trade-off, covering all prior task-specific distributions without hindering new task learning.
A hard adapter snapshot outperforms an EMA teacher, indicating that the exact post-training adapter state is a more reliable reference.
Alternative designs in computing the loss lead to suboptimal results, such as using CLS tokens from every block (rather than just the last block), all tokens from the last block, and using cosine distance (1-cosine similarity) in Eq.~\ref{eq:distill_loss} instead of MSE.


\begin{wraptable}{r}{0.4\columnwidth}
  \centering
  \vspace{-0.75em}
  \caption{Ablation on context prediction.}
  \scalebox{0.825}{
  \begin{tabular}{lcc}
    \toprule
    Predictor & $\mathbf{\bar{A}}$ & $\mathbf{A}_T$ \\
    \midrule
    Final classifier highest logits                    & 84.07 & 79.33 \\
    Aux. classifier highest logits                  & 84.19 & 78.63  \\
    Entropy as task context                             & 84.23 & 77.85 \\
    \rowcolor[rgb]{0.8, 0.9, 0.95}Min entropy & 84.70 & 79.97 \\
    Oracle                            & - & 81.62 \\
    \bottomrule
  \end{tabular}}
  \label{tab:abl_inference_predictor}
  \vspace{-0.75em}
\end{wraptable}

\paragraph{Task Context Prediction.}
Table~\ref{tab:abl_inference_predictor} investigates alternative methods to predict task context during inference.
Assembling the context vector using the highest logit in each task from the final classification head results in sub-optimal performance, while a context with the highest logits from auxiliary classifiers sees a 1.34\% drop in $A_T$, demonstrating the advantage of entropy for task identity prediction.
Directly filling the context using the entropy value for each task leads to a larger decline of $A_T$ to 77.85\%, confirming that the use of sparse context vectors is beneficial.
Notably, using ground truth task context during inference improves final accuracy to 81.62\%, implying that FACET can benefit from more accurate task context predictors for short task sequences. Additional analysis on task context prediction and robustness of FACET to task context prediction errors are given in the Appendix.

\section{Limitations}
FACET is designed for exemplar-free class-incremental learning with a frozen pretrained vision backbone, disjoint task label sets, and known task boundaries during training. These assumptions make the method directly comparable to PTM-based CIL methods, but they also leave several directions open. 
While FACET substantially reduces the task-specific costs compared to state of the art methods for long task sequences with 200 tasks, task-specific cost is kept small in the auxiliary classifiers but does not fully disappear.
Moreover, the single adapter capacity may saturate. Although we achieve strong performance up to 200 tasks, longer task sequences or highly heterogeneous tasks may eventually exceed what one adapter can represent. Extending our method to true lifelong continual learning over thousands of tasks is non-trivial.
In addition, the replay-free objective preserves old task-specific feature distributions by matching a frozen snapshot on current-task inputs, which is efficient and avoids storing old data. However, it is not equivalent to matching the full old data distribution and may be weaker when the data distribution of the new task is highly different from earlier tasks.
Promising directions include extending the method to large language models and studying alternative task context prediction methods.

\section{Conclusion}
This work has introduced FACET, a single-adapter approach for pretrained-model-based class-incremental learning.
By transforming the adapter's latent representation in a task-conditioned manner while preserving previously learned task-specific feature distributions, FACET makes continual learning more efficient and scalable. 
Future work could extend the proposed method's principles of sharing weights across tasks with dynamic conditioned transformation to extract task-specific features to other domains, such as large language models, vision-language models, and other scenarios where efficient continual learning is useful.

\bibliographystyle{ieee_fullname}
\bibliography{reference}


\appendix

\section{Additional Dataset and Implementation Details}

\subsection{Datasets}

We evaluate FACET on five datasets, including ImageNet-R~\cite{hendrycks2021ImagenetR}, ImageNet-A~\cite{hendrycks2021ImagenetA}, CIFAR100~\cite{krizhevsky2009CIFAR100}, ObjectNet~\cite{barbu2019objectnet,zhou2025APER}
, and OmniBenchmark-1K~\cite{zhang2022omnibenchmark,lou2026CARE}. 
The datasets used in the main comparison are summarized below:

\textbf{ImageNet-R}~\cite{hendrycks2021ImagenetR} has 200 classes with 30000 images. It consists of ImageNet-compatible classes rendered in artistic and non-standard visual styles.

\textbf{ImageNet-A}~\cite{hendrycks2021ImagenetA} contains 200 classes with 7475 images. The dataset consists of naturally adversarial images with ImageNet-compatible categories.

\textbf{CIFAR100~\cite{krizhevsky2009CIFAR100}} is a standard image classification benchmark that contains 100 classes and 60000 images.

Following~\cite{wang2025tuna,sun2025mos,zhou2025APER}, the \textbf{ObjectNet}~\cite{barbu2019objectnet,zhou2025APER} dataset we refer to for CIL contains 200 classes with 33137 images. It a subset of the original ObjectNet~\cite{barbu2019objectnet} dataset that contains 50000 images of 313 distinct objects with different backgrounds, rotations, and imaging viewpoints.

\textbf{OmniBenchmark-1K}~\cite{zhang2022omnibenchmark,lou2026CARE} has 188569 images and 1000 classes with non-overlapping data with ImageNet-1k. It is a subset of the clean version of OmniBenchmark~\cite{zhang2022omnibenchmark} with at least 100 images per class.

\subsection{Implementation Details}
We use the PILOT~\cite{sun2025pilot} codebase to run our experiments. Following~\cite{wang2022L2P,wang2022dualprompt,zhou2024EASE,sun2025mos,zhou2025APER,wang2025tuna,jiang2509MIN,liang2024inflora,Wu2025SDLoRA}, our main experiments use the ViT-B/16 backbone pretrained on ImageNet-21k. 
FACET inserts one task-conditioned adapter shared across all tasks in each transformer block, with two 384-dimensional hidden layers. 
The linear projections for generating context vectors ($W_s, W_U, W_V$) are initialized to zeros for the first step of training.
The auxiliary classifier is a linear layer that is trained to predict class labels directly. 
A zero-mean Gaussian noise with a standard deviation of 0.3 is added to the context vector during training to improve its robustness against imperfect context predictions during inference.
Optimization uses SGD with a cosine learning-rate schedule, a weight decay of $5\times10^{-4}$, a batch size of 16, a learning rate of 0.04, and 20 training epochs. 
Classifier alignment following~\cite{wang2025tuna} is included during training. 
For the short task sequence benchmarks with 10 to 20 tasks, we ensemble the predictions when using task context and when disabling the adapter by setting task context to zero. For long task sequences, predictions use only the FACET probabilities from a single backbone pass.

\section{Additional Experiments}


\begin{table}[h]
  \centering
  \vspace{-0.75em}
  \caption{Performance comparison on ImageNet-R (Inc10) using ViT-B/16-IN21k backbone and 5 random seeds. Metric is $\bar{A}$ / $A_T$.}
  \scalebox{0.76}{
    \begin{tabular}{lcccccc}
    \toprule
    Method $\backslash$ Seed & 49 & 127 & 982 & 1523 & 2026 & Mean $\pm$ Std \\
    \midrule
    SD-LoRA  & 78.50 / 70.58 & 77.27 / 70.73 & 77.82 / 71.85 & 76.58 / 69.95 & 77.22 / 70.77 & 77.48 $\pm$ 0.72 / 70.78 $\pm$ 0.68 \\
    TUNA & 82.51 / 77.78 & 81.55 / 76.97 & 82.37 / 77.73 & 82.44 / 77.68 & 82.44 / 77.15 & 82.26 $\pm$ 0.40 / 77.46 $\pm$ 0.37  \\
    MIN & 84.41 / 78.90 & 84.46 / 77.80 & 85.22 / 79.40 & 84.38 / 78.10 & 83.65 / 77.43 & 84.42 $\pm$ 0.56 / 78.33 $\pm$ 0.81 \\
    \rowcolor[rgb]{0.8, 0.9, 0.95} FACET & \textbf{85.63 / 80.80} & \textbf{85.14 / 79.32} & \textbf{85.89 / 80.98} & \textbf{84.66 / 79.52} & \textbf{84.77 / 79.28}  & \textbf{85.22 $\pm$ 0.53 / 79.98 $\pm$ 0.84} \\
    \bottomrule
    \end{tabular}%
    }
  \label{tab:in21k_seed}
\end{table}

\begin{table}[h]
  \centering
  \vspace{-0.75em}
  \caption{Performance comparison on ImageNet-R (Inc10) using ViT-B/16-IN1k backbone and 5 random seeds. Metric is $\bar{A}$ / $A_T$.}
  \scalebox{0.76}{
    \begin{tabular}{lcccccc}
    \toprule
    Method $\backslash$ Seed & 49 & 127 & 982 & 1523 & 2026 & Mean $\pm$ Std  \\
    \midrule
    SD-LoRA  & 81.40 / 74.65 & 81.19 / 75.72 & 80.45 / 74.77 & 80.03 / 74.75 & 79.82 / 73.32  &  80.58 $\pm$ 0.70 / 74.64 $\pm$ 0.86\\
    TUNA & 83.98 / 78.53 & 83.37 / 77.97 & 84.08 / 78.78 & 83.36 / 78.31 & 83.45 / 78.22 & 83.65 $\pm$ 0.35 / 78.36 $\pm$ 0.31 \\
    MIN & 85.01 / 79.35 & 84.26 / 77.95 & 85.75 / 80.03 & 83.73 / 78.47 & 83.86 / 78.28 & 84.52 $\pm$ 0.85 / 78.82 $\pm$ 0.85 \\
    \rowcolor[rgb]{0.8, 0.9, 0.95} FACET & \textbf{86.05 }/ \textbf{80.87} & \textbf{84.82} / \textbf{79.90} & \textbf{86.65} / \textbf{80.88} & \textbf{86.68} / \textbf{80.67} & \textbf{85.50 }/ \textbf{80.07} & \textbf{85.94 $\pm$ 0.79} / \textbf{80.48 $\pm$ 0.46} \\
    \bottomrule
    \end{tabular}%
    }
  \label{tab:in1k_seed}
\end{table}

\subsection{Robustness to Task Order and Pretrained Backbone}
\label{appendix_robust}
To evaluate the robustness and generalizability of our approach, we shuffle the task order with 5 random seeds and experiment with two widely used pretrained backbones, ViT-B/16-IN21k and ViT-B/16-IN1k~\cite{dosovitskiy2020vit}. Tables~\ref{tab:in21k_seed}-\ref{tab:in1k_seed} present our results. FACET outperforms strong baselines in all task orders on both backbones. This indicates that the advantage of FACET is consistent rather than an artifact of a particular task order.


\subsection{Detailed Ablations on Single Adapter Design}
\label{appendix:single_adapter}

\begin{wraptable}{r}{0.48\columnwidth}
  \centering
  \caption{Impact of initialization and noise on ImageNet-R (Inc10).}
  \scalebox{0.825}{
    \begin{tabular}{lcc}
    \toprule
    Config. & $\mathbf{\bar{A}}$ & $\mathbf{A}_T$ \\
    \midrule
    Random, no noise   & 82.74 & 76.68 \\
    Zero,   no noise   & 83.95 & 78.80 \\
    \rowcolor[rgb]{0.8, 0.9, 0.95}Zero, $+$noise   & 84.70 & 79.97 \\
    \bottomrule
    \end{tabular}%
    }
  \label{tab:abl_dcom_init}
\end{wraptable}

Random initialization starts each new task with a random transformation of the adapter feature space and reduces $A_T$ to 76.68\%.
Zero initialization avoids this unstable start by making a newly activated task-conditioned mixture component initially fall back to the pretrained representation, improving $A_T$ to 78.80\%.
Adding noise to the context vector further improves $A_T$ to 79.97\%, supporting its role in making the model robust to imperfect task identity prediction during inference.

Finally, Table~\ref{tab:abl_dcom_capacity} varies the bottleneck dimension and adapter depth.
The results are stable across a range of capacities, and deeper adapters do not consistently improve performance, indicating that FACET's gains are not simply due to adding more adapter parameters.

\begin{wraptable}{r}{0.5\columnwidth}
  \centering
  \vspace{-0.75em}
  \caption{Effect of $K_{max}$ on ImageNet-R (Inc10).}
  \scalebox{0.825}{
    \begin{tabular}{lcccc}
    \toprule
    Metric $\backslash$ Dim & 50 & 100 & \cellcolor[rgb]{0.8, 0.9, 0.95}200 & 400 \\
    \midrule
    $\mathbf{A_T}$ & 80.20 & 80.14 & \cellcolor[rgb]{0.8, 0.9, 0.95}79.97 & 79.64 \\
    $\mathbf{\bar{A}}$ & 84.77 & 84.73 & \cellcolor[rgb]{0.8, 0.9, 0.95}84.70 & 84.44 \\

    \bottomrule
    \end{tabular}%
    }
  \label{tab:abl_kmax}
\end{wraptable}

As shown in Table~\ref{tab:abl_kmax}, FACET is robust to different values of $K_{max}$, with smaller values achieving better performance for ImageNet-R (Inc10) with 20 tasks. We select 200 since it is a sufficiently large upper bound on the number of tasks in CIL.

\begin{table}[hb]
  \centering
  \vspace{-0.75em}
  \caption{Ablation of adapter capacity on ImageNet-R (Inc10). Metric is ($\mathbf{A}_T$).}
  \scalebox{0.825}{
    \begin{tabular}{lcccc}
    \toprule
    Hidden depth $\backslash$ Dim & 16 & 64 & 256 & 384 \\
    \midrule
    1 & 79.08 & 79.10 & 79.38 & 79.49 \\
    \rowcolor[rgb]{0.8, 0.9, 0.95}2 & 79.39 & 79.30 & 79.57 & \textbf{79.97} \\
    3 & 79.37 & 78.97 & 79.62 & 79.15 \\
    \bottomrule
    \end{tabular}%
    }
  \label{tab:abl_dcom_capacity}
\end{table}

\begin{table}[ht]
  \centering
  \vspace{-0.75em}
  \caption{Impact of the consistency weight $\lambda_{\mathrm{con}}$ on ImageNet-R (Inc10).}
  \scalebox{0.825}{%
  \begin{tabular}{lccccc}
    \toprule
    Metric & 0.5 & 1.0 & 2.0 & 4.0 & 8.0 \\
    \midrule
    $\mathbf{\bar{A}}$ & 83.94 & 84.75 & \cellcolor[rgb]{0.8, 0.9, 0.95}84.70 & 84.34 & 84.13 \\
    $\mathbf{A}_T$     & 77.12 & 79.28 & \cellcolor[rgb]{0.8, 0.9, 0.95}79.97 & 78.72 & 78.50 \\
    \bottomrule
  \end{tabular}%
  }
  \label{tab:abl_distill_weight}
\end{table}

\subsection{Impact of the Coefficient for Task-Dependent Feature Consistency Loss}

Table~\ref{tab:abl_distill_weight} shows that  $\lambda_{\mathrm{con}}=2.0$ achieves the highest final accuracy, achieving an optimal balance between mitigating forgetting of task-specific feature distributions and learning new task.




\section{Analysis of Task Prediction Accuracy}
We further evaluate the robustness of FACET to wrong task-context predictions during inference.
Without retraining, we compare with a randomly assigned context and a deliberately incorrect context on the same checkpoint.
Random and forced-wrong results are mean $\pm$ sample standard deviation over five tries.

\begin{table}[htbp]
\centering
\caption{Robustness of FACET to wrong task-context predictions at inference (same checkpoint, no retraining).
Random and forced-wrong results are mean $\pm$ sample standard deviation over five tries.}
\label{tab:facet-wrong-context}
\resizebox{\linewidth}{!}{%
\begin{tabular}{@{}lccccc@{}}
\toprule
Setting &
\begin{tabular}[c]{@{}c@{}}Task-context pred.\ acc.\\(random choice)\end{tabular} &
\begin{tabular}[c]{@{}c@{}}Final acc.\\w/ predicted ctx.\end{tabular} &
\begin{tabular}[c]{@{}c@{}}Final acc.\\w/ oracle ctx.\end{tabular} &
\begin{tabular}[c]{@{}c@{}}Final acc.\\w/ random ctx.\end{tabular} &
\begin{tabular}[c]{@{}c@{}}Final acc.\\w/ forced-wrong ctx.\end{tabular} \\
\midrule
ImageNet-R Inc20 & 53.83\% (10\%) & 80.66 & 86.38 & $78.07 \pm 0.19$ & $77.54 \pm 0.15$ \\
ImageNet-R Inc10 & 49.05\% (5\%)  & 79.97 & 81.62 & $75.80 \pm 0.24$ & $75.49 \pm 0.28$ \\
ImageNet-R Inc4  & 15.21\% (2\%)  & 76.63 & 78.72 & $76.59 \pm 0.13$ & $76.47 \pm 0.09$ \\
\bottomrule
\end{tabular}%
}
\end{table}

Task prediction accuracy is $5.3$ to $9.8$ times above random guesses.
More importantly, using an incorrect context reduces final accuracy by at most $4.48$ ($79.97$ to $75.49$ for ImageNet-R 20 tasks).
We note that $75.49$ is comparable to the performance of the strong baseline SD-LoRA ($75.26$).
This shows that FACET is robust to wrong task-context predictions during inference.
We note that providing ground-truth task context during inference is beneficial (ablation Table~\ref{tab:abl_inference_predictor}), and can improve final accuracy from $79.97$ to $81.62$ on ImageNet-R 20 tasks.
Therefore, the performance of FACET is not only positively correlated with task-context prediction accuracy, but also robust against wrong task-context predictions.
More advanced mechanisms to predict task context can further improve performance.

\section{Additional Analysis of Task-Conditioned Feature Consistency}
\label{app:snapshot}

\begin{figure}[hbt]
    \centering
    \includegraphics[width=0.45\textwidth]{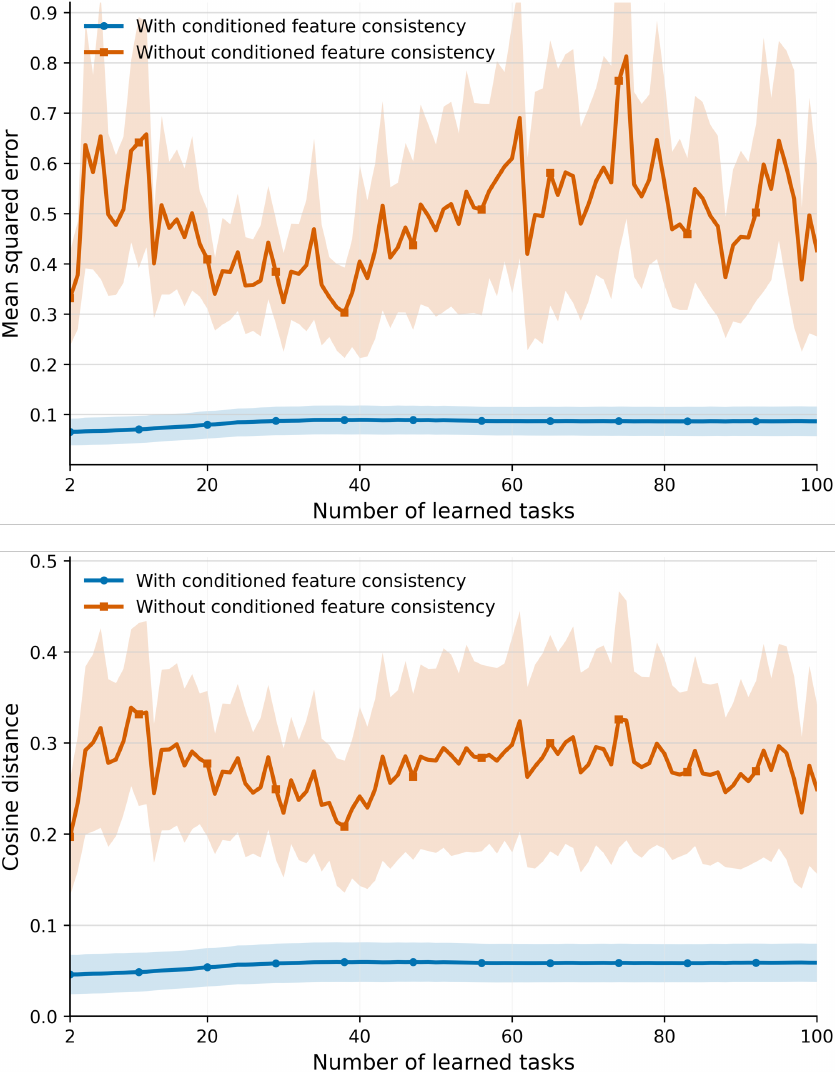}
    \caption{How well our single adapter preserves the feature distribution of the first task as it is sequentially trained on new tasks from task 2 to 100. \textbf{Top}: Mean square error as the distance metric. \textbf{Bottom}: Cosine distance as the distance metric. We compute the distance between the CLS feature produced by the model after it has learned the first task with the corresponding CLS feature after the model has sequentially learned $t \in \{2,...,100\}$ tasks. The original training data and context for the first task are used for a direct comparison.}
\label{fig:feature_drift}
\end{figure}

Here we evaluate how well our conditioned feature consistency loss preserves the feature distributions of historical tasks. Specifically, we measure how well the sample features from the first task are preserved after learning new tasks.
The CLS token from the last transformer block is selected as the feature representation since it is passed to the classifier for prediction. 
Note that the task-conditioned feature consistency loss is computed using data from task $t$ ($x\in \mathcal{D}_t$), which is the latest task the adapter is being trained on.

Results are shown in Fig.~\ref{fig:feature_drift}, where the solid line denotes the mean distance and the light shaded area illustrates the standard deviation. It can be seen that our conditioned feature consistency loss (blue line) effectively preserves the original features from the first task, even after the adapter has been trained on 99 subsequent tasks from the OmniBenchmark-1K dataset. 
Without our conditioned feature consistency loss (orange line), the MSE and Cosine distances are much larger, which indicates forgetting of the task-specific distribution.
For instance, after task 100, the mean cosine distance and mean squared error when using the consistency loss are 0.0588 and 0.0866, respectively. Without the consistency loss, cosine distance and MSE increases to 0.249 (+323\%) and 0.427 (+393\%), respectively.
These results confirm the effectiveness of the conditioned feature consistency loss in preserving task-specific feature distributions in continual learning even though it relies on data from the latest task instead of those from the original tasks.

We provide more details on how cosine distance and MSE are computed below:
We first extract the CLS features of the data samples from the first task using the adapter trained on its dataset $\mathcal{D}_1$. 
This defines the original feature distribution of the first task. 
Then, after learning $t$ tasks, we extract CLS features using the same data samples from $\mathcal{D}_1$ and the same context vector for the first task, but using the adapter sequentially trained on all $t$ tasks.
We evaluate $t \in \{2,...,100\}$ using the OmniBenchmark-1K dataset.
Because the input images and context vectors are the same, any difference between corresponding CLS features directly measures how much an original feature from the first task has drifted after learning later tasks.

We quantify this drift using cosine distance and mean squared error (MSE). For each training image $x\in \mathcal{D}_1$, cosine distance is computed between the original CLS feature and the corresponding CLS feature extracted after learning $t$ tasks:
\[
    d_{\cos}
    =
    1 -
    \frac{
        \langle \mathrm{CLS}^{\theta_1}_{1}, \mathrm{CLS}^{\theta_t}_{1} \rangle
    }{
        \|\mathrm{CLS}^{\theta_1}_{1}\|_2
        \|\mathrm{CLS}^{\theta_t}_{1}\|_2
    },
\]
where $\mathrm{CLS}^{\theta_1}_{1}$ is the original CLS feature from the model trained on the first task, and $\mathrm{CLS}^{\theta_t}_{1}$ is the CLS feature from the model after learning $t$ tasks using the same context. We average the cosine distance over all data samples.
We also compute the MSE distance:
\[
    d_{\mathrm{MSE}}
    =
    \frac{1}{N}
    \left\|
        \mathrm{CLS}^{\theta_1}_{1}
        -
        \mathrm{CLS}^{\theta_t}_{1}
    \right\|_2^2,
\]
where $N$ is the number of images in the first task. These two metrics are complementary: cosine distance captures directional drift, while MSE captures overall feature-space displacement.

\begin{table}[h]
\centering
\caption{Task-1 forgetting versus CLS feature drift under different $\lambda_{\mathrm{con}}$ on ImageNet-R Inc10.}
\label{tab:feature-drift-forgetting}
\scalebox{0.825}{
\begin{tabular}{cccc}
\toprule
$\lambda_{\mathrm{con}}$ & Task-1 forgetting & Cosine distance & MSE \\
\midrule
0   & 35.96 & 0.2376 & 0.5072 \\
0.5 & 36.59 & 0.0888 & 0.1578 \\
1   & 31.23 & 0.0538 & 0.0961 \\
2   & 18.61 & 0.0285 & 0.0502 \\
4   & 15.46 & 0.0167 & 0.0296 \\
8   & 13.32 & 0.0113 & 0.0206 \\
\bottomrule
\end{tabular}
}
\end{table}

We also directly test whether feature drift is related to old-task classification forgetting.
On ImageNet-R Inc10, we train six models that differ only in $\lambda_{\mathrm{con}}$.
After task 20, we compare the CLS features of the task-1 images with their features immediately after task 1, always using the task-1 context, and measure task-1 forgetting.

Across the six settings, cosine distance and MSE have Pearson correlations of $0.752$ and $0.711$ with task-1 forgetting.
Greater drift is strongly associated with greater old-task classification forgetting.

\section{Justification of Using Current-Task Data for Conditioned Feature Consistency}
In Equation~\ref{eq:distill_loss}, we use the data from the current task $D_t$ to preserve the old feature distributions. Here we provide a justification on why it works.

Current-task images are probe inputs used to detect whether new-task training
changes the shared adapter under a previous context. They are not treated as
approximations of old-task images in the consistency loss.

After task \(t{-}1\), we freeze a copy of the adapter. While learning task
\(t\), a current image is processed by the frozen and updated adapters under
the same sampled old context \(c_j\). Because the input and context are
identical, their feature difference isolates the effect of the adapter
update. Minimizing this difference therefore constrains new-task training from
changing the adapter's behavior under \(c_j\).

The backbone is frozen and all tasks use the same low-dimensional adapter.
Consequently, current probes can protect old features when the same adapter
weights affect both current and old features. In that case, a weight change
that damages old features also changes the current probe features and is
penalized by the consistency loss. This does not require the tasks to have
identical images, labels, or feature distributions.

\paragraph{Local Justification.}
Let \(\Delta\theta=\theta-\theta^-\) be one adapter update and let
\(h_\theta(x,c_j)\) be the conditioned feature. For a small update,
\begin{equation}
h_\theta(x,c_j)-h_{\theta^-}(x,c_j)
\approx J_j(x)\Delta\theta,
\end{equation}
where \(J_j(x)\) is the feature Jacobian with respect to the shared-adapter
parameters. The current-input consistency loss and old-input feature drift are
locally
\begin{equation}
\mathcal{L}_{\mathrm{con}}^{j}
\approx \Delta\theta^\top G_t^j\Delta\theta,
\qquad
D_j\approx\Delta\theta^\top G_j^j\Delta\theta,
\end{equation}
where
\begin{equation}
G_a^j
=\mathbb{E}_{x\sim\mathcal{D}_a}
\bigl[J_j(x)^\top J_j(x)\bigr].
\end{equation}
If every small adapter change that affects old features is also detected by
the current probes, up to a constant \(\kappa\),
\begin{equation}
G_j^j\preceq\kappa G_t^j,
\end{equation}
then
\begin{equation}
D_j\leq\kappa\mathcal{L}_{\mathrm{con}}^{j}.
\end{equation}
Thus, minimizing consistency on current inputs also limits old-task feature
drift when current and old inputs overlap and respond to the same adapter
changes.

\paragraph{Empirical justification for sufficient overlap.}
Overlap means that changing the same adapter weights affects both sets of
conditioned features in similar ways. It does not require current and old
images, or even their frozen-backbone features, to be similar one by one.
\(J_j(x)\) records the effect of small adapter-weight changes for one input,
and \(G_t^j\) and \(G_j^j\) summarize these effects over current and old
inputs.

We test this overlap directly with feature-Jacobian products over all
shared-adapter parameters. We use ImageNet-R (Inc10) and analyze a checkpoint
obtained while training the second task. At this fixed checkpoint, we
construct two input sets. The first contains old inputs, namely images from
the first task. The second contains current probes, namely images from the
second task. Both sets are processed using the old-task context \(c_1\).
Therefore, the only difference between \(G_1^0\) and \(G_2^0\) is the input
distribution,
\begin{equation}
G_1^0
=\mathbb{E}_{x\sim\mathcal{D}_1}
\bigl[J_1(x)^\top J_1(x)\bigr],
\qquad
G_2^0
=\mathbb{E}_{x\sim\mathcal{D}_2}
\bigl[J_2(x)^\top J_2(x)\bigr].
\end{equation}
For each image set, we measure how the \(768\)-dimensional conditioned feature
responds to all shared-adapter parameters. The auxiliary classification heads
are excluded because they do not produce the conditioned feature matched by
the consistency loss.

\begin{table}[t]
\centering
\caption{Feature-Jacobian overlap between current probes and old inputs under
the old-task context \(c_1\) on ImageNet-R (Inc10).}
\label{tab:jacobian_overlap}
\small
\setlength{\tabcolsep}{4pt}
\begin{tabular}{@{}p{0.30\linewidth}c p{0.52\linewidth}@{}}
\toprule
Measurement & Result & Interpretation \\
\midrule
Similarity between adapter changes detected by current and old inputs
& \(0.809{\pm}0.041\)
& Normalized similarity between \(G_1^0\) and \(G_2^0\).
A value of \(1\) means identical sensitivity structure up to scale, and
\(0\) means orthogonal structure. The result indicates substantial overlap. \\
Rank-\(16\) old-gradient energy captured by current directions
& \(85.2{\pm}3.5\%\)
& We learn the \(16\) strongest adapter directions from current probes and
test them on held-out old inputs. They capture \(85.2\%\) as much old-input
Jacobian variance as directions learned directly from old inputs. \\
Sampled old-active directions satisfying
\(v^\top G_1^0 v \leq 2\, v^\top G_2^0 v\)
& \(96.88\%\)
& For \(96.88\%\) of sampled directions that affect old features, the
old-input effect is no more than twice the effect detected by current
probes. \\
Sampled old-active directions satisfying
\(v^\top G_1^0 v \leq 5\, v^\top G_2^0 v\)
& \(100\%\)
& Every sampled old-active direction is detected by current probes within a
factor of \(5\). \\
\bottomrule
\end{tabular}
\end{table}

These measurements support the required coverage interpretation.
Current-task images, when evaluated under the old-task context, activate most
adapter directions that are important for first-task features. The experiment
does not claim that old- and current-task images are semantically similar.

We verified that this behavior remains after the full \(20\)-task sequence.
At the final checkpoint, we use images from the last task as current probes
and compare them with images from old tasks \(0\), \(4\), \(8\), \(12\), and
\(16\). For each old task \(j\), both image sets are evaluated under that
task's context \(c_j\). Across these five comparisons, the average similarity
between current-task directions and old-task directions is
\(0.869{\pm}0.049\), and \(99.58\%\) of sampled old-active directions satisfy
the directional relation with \(\kappa{=}5\).

\end{document}